\documentclass{article}

    \usepackage[final,nonatbib]{neurips_2024}

\usepackage[utf8]{inputenc} % allow utf-8 input
\usepackage[T1]{fontenc}    % use 8-bit T1 fonts
\usepackage{hyperref}       % hyperlinks
\usepackage{url}            % simple URL typesetting
\usepackage{booktabs}       % professional-quality tables
\usepackage{amsfonts}       % blackboard math symbols
\usepackage{nicefrac}       % compact symbols for 1/2, etc.
\usepackage{microtype}      % microtypography
\usepackage{xcolor}         % colors
\usepackage{amsmath}
\usepackage{graphicx}
\usepackage{algorithm}
\usepackage{algorithmic}
\usepackage{wrapfig}
\usepackage{booktabs} % For \toprule, \midrule, and \bottomrule
\usepackage{makecell} % For \makecell if you are using it
\usepackage{multirow} % For multirow

\newtheorem{definition}{Definition}

\title{LLM-Assisted Red Teaming of Diffusion Models through ``Failures Are Fated, But Can Be Faded''}

\author{%
  Som Sagar \\
  Arizona State University \\
  \texttt{ssagar6@asu.edu} \\
  \And
  Aditya Taparia \\
  Arizona State University \\
  \texttt{ataparia@asu.edu} \\
  \And
  Ransalu Senanayake \\
  Arizona State University \\
  \texttt{ransalu@asu.edu} \\
}

\begin{document}

\maketitle

\begin{abstract}
In large deep neural networks that seem to perform surprisingly well on many tasks, we also observe a few failures related to accuracy, social biases, and alignment with human values, among others. Therefore, before deploying these models, it is crucial to characterize this failure landscape for engineers to debug or audit models. Nevertheless, it is infeasible to exhaustively test for all possible combinations of factors that could lead to a model's failure. In this workshop paper, we improve the ``Failures are fated, but can be faded'' framework~\cite{Sagar2024icml}---a post-hoc method to explore and construct the failure landscape in pre-trained generative models---with a variety of \emph{deep reinforcement learning} algorithms, screening tests, and LLM-based rewards and state generation. With the aid of limited human feedback, we then demonstrate how to restructure the failure landscape to be more desirable by moving away from the discovered failure modes. We empirically demonstrate the effectiveness of the proposed method on diffusion models. We also highlight the strengths and weaknesses of each algorithm in identifying failure modes.
\end{abstract}

\section{Introduction}

\begin{figure*}[t]
\begin{center}
\centerline{\includegraphics[width=1\columnwidth]{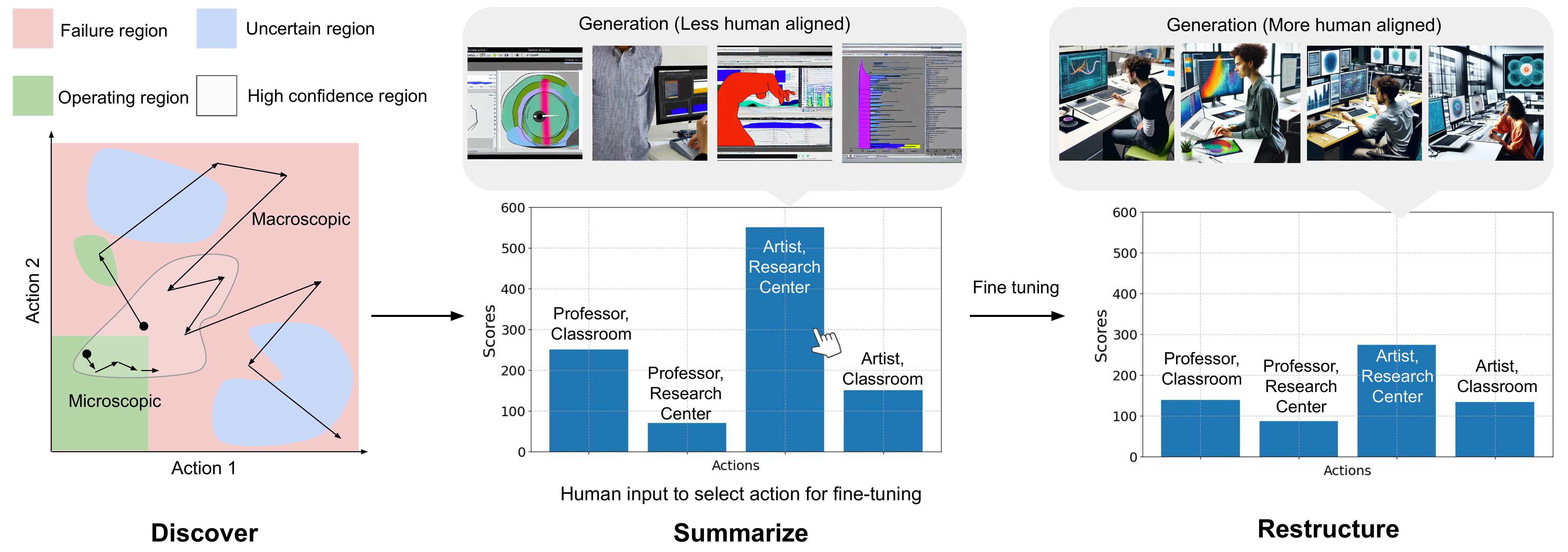}}
\caption{There are three main steps in the proposed failure discovery and mitigation framework. \textbf{1. Discover}: We propose a deep RL-based method to explore the \emph{failure landscape} with microscopic and macroscopic exploration strategies. It will discover regions where the model works and fails, with varying levels of confidence. \textbf{2. Summarize}: Results are qualitatively and quantitatively summarized for the user to indicate preferences. \textbf{3. Restructure}: Based on the user's preferences from the previous stage, the model can be fine-tuned to mitigate or shift away the failure modes to unlikely regions. The center image shows images generated by Stable Diffusion v1-4 for the prompt ``\textit{Create an image of a distinctive \textless artist\textgreater ~analyzing data on a computer in a \textless research center\textgreater}''. A user selects the most likely failure in terms of image quality from the summary report. The fine-tuned model, based on user preferences, has generated more naturalistic images.}
\label{fig:failure_landscape}
\end{center}
\vskip -0.2in
\end{figure*}

No dataset or model, regardless of its size, can encompass the full spectrum of real-world scenarios. Consequently, they are expected to fail under certain conditions. However, unlike in white-box modeling, where we construct models from first principles by clearly defining assumptions, it is impossible to know \textit{a priori} which factors contribute to the failures of deep learning models. These failures often only become apparent after deployment, when the models are exposed to diverse and unpredictable real-world data. To name a few examples of failures: commercial generative AI-based platforms that are susceptible to producing stereotypical or racist outputs can cause societal stigma and perpetuate bias. The importance of identifying such failure modes stems from two different aspects. First, engineers and data scientists need to understand the numerous factors that affect model performance to debug these models. Second, policymakers, legislative bodies, and insurance companies need an accessible method to audit the capabilities of these models. As illustrated in Fig.~\ref{fig:failure_landscape}, the main requirement for both stakeholders is an efficient tool that can automatically explore various areas of the failure landscape.

Although users of deep neural network-based systems frequently encounter failures, as evidenced by daily social media posts, there have been relatively few attempts to develop techniques for exploring the landscape of these failures. This is primarily due to the exceedingly high number of test cases, rendering classical search techniques impractical. Models often fail due to combinations of factors in the continuous, discrete, or hybrid domains. A model might fail in one case while performing adequately in another seemingly similar case, emphasizing the stochastic nature of the failure landscape and thus exacerbating the difficulty of the problem. For instance, as shown in the histogram of Fig.~\ref{fig:failure_landscape}, changing the profession in a text prompt result in bias.

To tackle these challenges, we need a method that can explore large spaces by taking many possible actions while also taking into account the stochasticity of the system. As a solution, we propose a deep Reinforcement Learning (deep RL)-based method to post-hoc characterize the failure landscape of large-scale pre-trained deep neural networks. The deep RL-based algorithm iteratively interacts with the environment (i.e., the model we want to audit) to learn a stochastic policy that can find failures by satisfying criteria, either implicit or explicit, provided by a human. We propose various operating modes of the deep RL-based algorithm to explore the failure landscape with different specificities as engineers and legislative bodies have different needs.

% \begin{wrapfigure}{r}{0.38\textwidth}  
% \vspace{-1em}
%     \centering
%     \includegraphics[width=\linewidth]{images/Adversarial training.pdf}
%     \vskip -0.1in
%     \caption{Visualization of failure landscape and search on adversarially trained model.}
%     \label{fig:adversarial_training}
% \end{wrapfigure}

Characterizing the failure landscape is not useful if it cannot be used to improve the model. By taking a limited amount of human feedback, we show how the harmful and frequent failures can be mitigated, showing the effectiveness of our failure detection and representation mechanisms. In this workshop paper, we make several improvements to ``Failures are Fated, But Can be Faded''~\cite{Sagar2024icml}:
\begin{enumerate}
    \item A method to automate reward and state collection through LLMs is devised.
    \item In addition to DQN, the framework is generalized to add other RL algorithms such as PPO and A2C. We also show how different RL algorithms explores failure modes differently. 
    \item Inspired by design of experiments (DOE)~\cite{montgomery2017design}, a screening mechanism to reduce the action space is introduced. 
\end{enumerate}

\section{Related Work}
\label{sec:related}

{\bf Formal verification and validation} of neural networks is an active field of research~~\cite{huang2017safety}. Statistical approaches have also been used for verifying neural networks~\cite{bartlett2021deep}. While the advances in these fields are important, in its current state, these approaches struggle with scaling to SOTA deep neural networks. Therefore, considering the rapid deployment of these models, taking a completely empirical approach, we develop alternative techniques to characterize the failure landscape. Accurate failure categorization is crucial for understanding model limitations as demonstrated by methods such as those applied in deep regression models~\cite{thiagarajanpager}.

{\bf Out-of-distribution (OOD)} detection research aims to determine whether a given input is OOD~\cite{fort2021exploring,nitsch2021out,anirudh2023out}. In most cases, it is challenging to discern whether the learned model is underperforming or the data is genuinely OOD. To address these challenges, recent methods, such as ~\cite{subramanyam2024decider} leverages LLMs and VLMs to detect failures by aligning visual features to core attributes, ~\cite{thiagarajan2022single} estimate uncertainty by accounting for prediction inconsistencies under biased data. Rather than focusing on detecting OOD inputs, our work emphasizes identifying regions where failures are likely to occur.

{\bf Adversarial attacks}~\cite{madry2017towards,silva2020opportunities} can be thought as a way to make data points OOD by applying a small perturbation. They, if necessary, can be categorized as a sub-case of our exploration around the origin of the concept space. However, this paper, specifically looks at characterizing the whole failure landscape of interest, rather than the sensitivity to small perturbations. This complete characterization is more actionable, providing an interface for the engineers to debug models or auditing bodies to understand limits. 

{\bf Reinforcement Learning} We explore three popular RL algorithm Deep Q-Network (DQN)~\cite{mnih2015humanlevel}, Proximal Policy Optimization 
 (PPO)~\cite{schulman2017proximal}, Advantage Actor-Critic (A2C)~\cite{pmlr-v48-mniha16}. DQN combines Q-learning with deep neural networks to approximate Q-values for each state-action pair. PPO is a policy gradient method that balances between policy improvement and stability. It employs a clipped surrogate objective to prevent large policy updates, leading to more stable and efficient training. A2C is a synchronous version of the actor-critic method that estimates the advantage function to reduce variance in policy updates. The actor learns the policy, while the critic evaluates the current policy by learning a value function. RL has been shown to be an effective search strategy in a wide range of applications, such as drug discovery~\cite{popova2018deep} etc. Previously, MDPs with solvers such as Monte Carlo Tree Search have been applied to perturb individual LIDAR points or pixels~\cite{Delecki2022arxiv} and states of aircraft and autonomous vehicles~\cite{corso2021survey}. Such techniques, while ideal for the applications considered, become quickly infeasible in high-dimensional continuous action spaces as in testing foundation models. Further, since such data-driven stress testing methods in aeronautics engineering can be formulated as reinforcement learning-based adversarial attacks in machine learning~\cite{yang2020patchattack,wang2021reinforcement}, limitations of adversarial attacks still hold. Rather than devising methods for adversarial attacks, our aim is to red team and characterize the whole failure landscape to subsequently mitigate them. Most work on red teaming~\cite{eyuboglu2022domino, ganguli2022red, jain2022distilling, prabhu2024lance}, except~\cite{Sagar2024icml, hong2024curiositydriven}, do not pose failure discovery as a reinforcement learning problem. ~\cite{hong2024curiositydriven} is limited to LLMs, whereas we improve the usability of the text-to-image generation in~\cite{Sagar2024icml} through LLMs. The scalability of the proposed method is primarily attributed to the capabilities of deep RL to manage large and high-dimensional action spaces effectively.

\section{Characterizing the Failure Landscape}
\label{sec:methods}
\subsection{Defining Failures}

Let us consider a deep neural network\footnote{A generative model $f: Z \rightarrow X^\prime$ is a mapping from the space of learned latent variables, $z \in Z$, to the space of generated data, $y \in X^\prime$. To keep the subsequent discussion clearer, we have intentionally abused notation here by reusing and overloading $f$ and $y$. Therefore, intuitively, $y$ is the output of the network during inference.} $f_\theta$, parameterized by $\theta$, produces an output $y$. Like any model, $f_\theta$ operates only under certain conditions, although these conditions are not evident for deep neural networks. Even if we can find all the valid operating conditions, merely enumerating them is not sufficient to address the model's issues. Therefore, our goal is to identify a set of specific operating conditions, which we refer to as concepts $C$, under which the model $f_\theta$ is most likely to fail. 

\begin{definition}
Let $m(.)$ be a scoring function that evaluates an output of a neural network. The discrepancy $\Delta$, under concepts $C$, is defined as the difference between the score of the human-specified output $m(y_\text{human})$ and the score of the model's output $m(y)$. The model is considered to have failed under $C$, if $\Delta(m(y_\text{human}),m(y)|C) > \epsilon$, for some non-negative $\epsilon$.
 \label{def:1}
\end{definition} 

Here, $y_\text{human}$ can be annotated ground truth labels or run-time human evaluations~\cite{christiano2017deep}. Therefore, $y_\text{human}$ indicates human's expectation on what the output should be. The discrepancy $\Delta$ can simply be a scoring scheme used in generative AI image evaluation. For example, in the case of a text-conditioned image generation task, $y_\text{human}$ can be a combination of image quality, gender bias, and art style, while $C$ can be a combination of profession-related terms and grammatical mistakes in the text prompt. Certain combinations of $C$, results in larger $\Delta$. Since discovering all inputs that lead to failures under $C$ is neither feasible nor useful, our objective is to craft an algorithm to efficiently modify these concepts to adequately explore the failure landscape.  

\subsection{Discovering Failures}

Our objective is to modify concepts $C$ in such a way that the model fails. To handle the stochasticity of the input-output mapping and large continuous or discrete concept set for large datasets, we frame this as a deep RL problem. We want to find a policy $\pi$ that can alter the values of these concepts by applying actions $a$ on concepts $C$. For instance, if $C=\{\text{gender}=\{\text{male, female}\}, \text{profession}=\{\text{professor, musician, chef}\}\}$, actions for a prompt \textit{Generate a \textless gender\textgreater \textless profession\textgreater} under $C$, will consider different combinations of $C.$ An example of an action is \textit{Generate a male chef}.

To learn the policy that can suggest the best actions, we consider a Markov Decision Process (MDP), defined by the tuple \( (S, A, P, R, \gamma) \), for set of states (observation space) \( S \), set of actions (action space) \( A \), a transition probability function \( P: S \times A \times S \rightarrow [0, 1] \), reward function \( R: S \times A \rightarrow \mathbb{R} \), and a discount factor \( \gamma \in [0, 1] \). An agent in state $s \in S$ takes the action $a \in A$ and transition to the next state $s^\prime \in S$ with transition probability $P(s^\prime|s,a)$. In other words, the RL algorithm samples a prompt $s$ and changes the value of the concept $c$ according to $a$, and obtain a new image or a prompt $s^\prime$, altered under $c$. By passing this new prompt through the neural network, we collect a reward $R(s^\prime,a)$. To encourage discovering failures, we define the reward function in such a way that the higher the probability of failure, the higher the $R$ is.

% Since the state and action spaces are large for the large-scale neural networks we consider, techniques such as vanilla Q learning~\cite{Sutton1998} are intractable. Therefore, we resort to Deep Q networks (DQNs)~\cite{mnih2015humanlevel}. DQNs process some additional attractive properties for characterizing the failure landscape: they can handle continuous actions spaces, generalize to unseen images and prompts, and remove correlation in sampling because of the replay buffer. DQN aims to learn an optimal policy \( \pi^* \) that maximizes the expected cumulative reward,
% \begin{equation}
%     Q^*(s, a) = \mathbb{E}_{s' \sim P(\cdot | s, a)}[R(s, a) + \gamma \max_{a' \in A} Q^*(s', a')].
% \end{equation}

\begin{wrapfigure}[17]{l}{0.51\textwidth} 
\vspace{-2em}

\begin{minipage}[b]{0.51\textwidth}
    \begin{algorithm}[H]
    \caption{Action Screening}
    \begin{algorithmic}[1]
    \STATE \textbf{Input:} Actions $A$, states $S$, reward function $R$
    \STATE \textbf{Output:} High-rewarding actions $A'$
    \STATE Initialize $R\_sums[a] \gets 0$ for each $a \in A$
    
    \FOR{each $s \in S$}
        \FOR{each $C \subseteq A$}
            \STATE $reward \gets R(C, s)$
            \FOR{each $a \in C$} 
                \STATE $R\_sums[a] \gets R\_sums[a] + reward$
            \ENDFOR
        \ENDFOR
    \ENDFOR
    
    \STATE $mean\_R \gets \frac{1}{|A|} \sum_{a \in A} R\_sums[a]$
    \STATE $A' \gets \{a \in A : R\_sums[a] \geq mean\_R\}$
    \STATE \textbf{return} $A'$
    \end{algorithmic}
    \label{alog:screen}
    \end{algorithm}
    
\end{minipage}
\end{wrapfigure}

% We employ the DQN algorithm with a fully-connected neural network as the policy. Since we want the DQN to initially explore the full landscape but later focus more on areas where failures are common, we set a learning rate schedule that gradually drops the exploration parameters from \( \epsilon_i = 1.0 \) to \( \epsilon_f = 0.6 \) over training episodes.

\textbf{Exploration:} We define the concept value set $C$ to contain all possible combinations of actions. The exploration is designed to caste a wide net to explore various areas quickly and identify regions of the action space where the model fails. These regions might be scattered across the space and not contiguous with the model's known operating region (i.e., the region where there are no failures). 

\textbf{Screening Experiments for the Factorial Design:} Since $C$ can increase exponentially with more actions, we perform a preliminary test to limit the selection of $a$ to high-rewarding actions. We evaluate all $C$ using the rewards obtained from randomly sampled states. We then assess each action's individual contribution to the rewards. If an action's contribution is below the mean average, we exclude it from consideration. In design of experiments~\cite{montgomery2017design} terms, we screen main effects before considering all interaction effects. The final results are shown in Appendix~\ref{appendix:prompts}

\subsection{The Deep RL Formulation with LLM-based States and Rewards}
\label{sec:ml_tasks}

We developed an RL environment based on the OpenAI's Gym library~\cite{gym}, focusing on text-to-image generation tasks.

\begin{figure*}[ht]
\begin{center}
{\includegraphics[width=0.45\columnwidth]{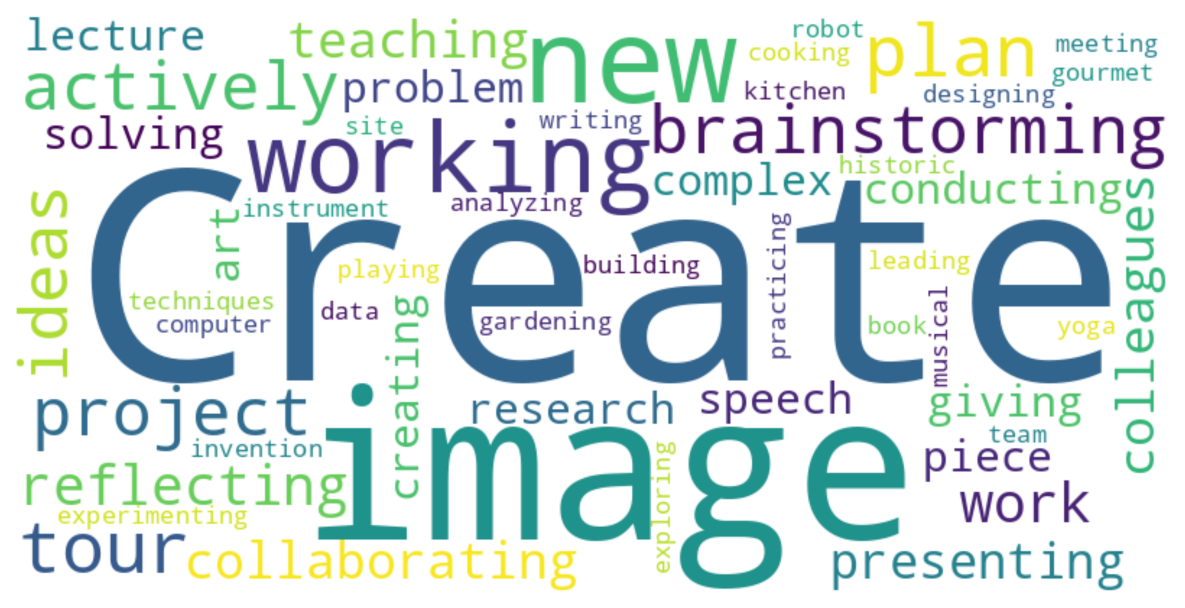}}
{\includegraphics[width=0.45\columnwidth]{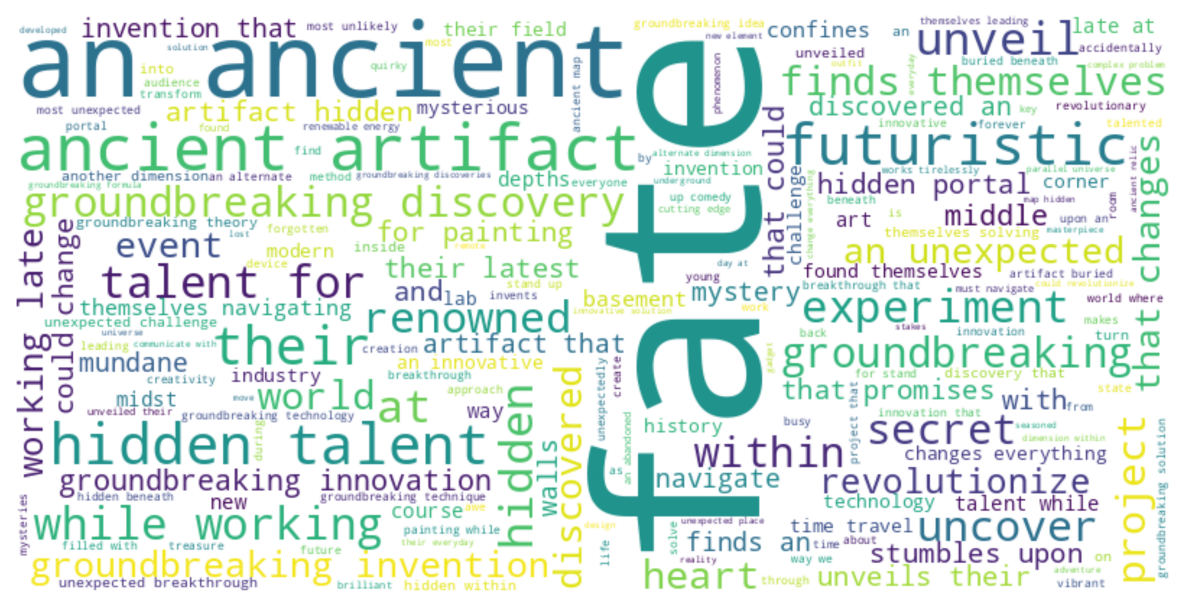}}
\caption{Wordcloud of prompts from a) predefined states b) LLM generated states}
\label{fig:llm_wordcloud}
\end{center}

\end{figure*}

\textit{Problem setup}: We utilize Stable Diffusion-v1-4 (SD v1-4)~\cite{Rombach_2022_CVPR}, a text-to-image model to generate images prompts, according to $C$. The action space consists of words from three sets of personal attribute, profession, and place (See Appendix~\ref{appendix:prompts}). These three categories were chosen because they represent key aspects of human identity and contextual scenarios that significantly influence the content and style of the generated images. By varying attributes (e.g., "unique" or "visionary"), profession (e.g., "mathematician" or "writer"), and place (e.g., "corporate office" or "research center"), we can systematically explore how different combinations affect the model's output, ensuring a wide range of scenarios are covered for testing failure modes.

\textit{State generation using an LLM}: The states are represented by prompts that are dynamically generated to simulate a wide range of scenarios for the diffusion model to process, this allows for scalability, as the generation of a vast number of prompts can be achieved rapidly with minimal manual effort, making it a highly efficient method for state generation in RL tasks. In contrast, \cite{Sagar2024icml} relied on predefined prompts faced the limitation of a fixed and narrow set of scenarios, which restricted exploration of diverse failure modes. Using LLM for state generation produced 1,310 unique words, whereas using 21 self-defined prompts resulted in only 61 unique words, as shown in the form of word-clouds in Fig.~\ref{fig:llm_wordcloud}. We achieve this by utilizing the GPT-4o language model to create prompts containing specific placeholders. The RL agent interacts with these states by selecting actions (such as choosing specific words to fill the placeholders), which are designed to explore potential failure modes of the diffusion model. The use of placeholders and GPT-4 allows for the rapid generation of a large number of diverse prompts without manual crafting.  

 The agent selects an action from the combination of attributes, professions, and  places (Total number of combinations depends on the number of action we get after the screening) and combines it with a base prompt from the observation space, and pass it through SD v1-4 to generate the image. For example, if the agent returns the \textless attribute\textgreater~to be unique, \textless profession\textgreater~to be scientist and \textless place\textgreater~to be corporate office, then the a final prompt example can be: ``\textit{Create an image of a \underline{unique} \underline{scientist} brainstorming new ideas in a \underline{corporate office}}.''

\textit{RL Agent}:
The RL agent learns to identify which combination of words from attributes, professions, and places results in worst image quality and has the most bias based on the given prompt from a LLM based reward function inspired by the success of designing reward function using LLM~\cite{yu2023language, kwon2023reward} and code-as-policy methods in procedural content generation, where RL is used to optimize code actions for task performance~\cite{liang2023code}.

\begin{wrapfigure}[15]{r}{0.38\textwidth}
\vspace{-2.5em}
    \centering
    \includegraphics[width=0.38\textwidth]{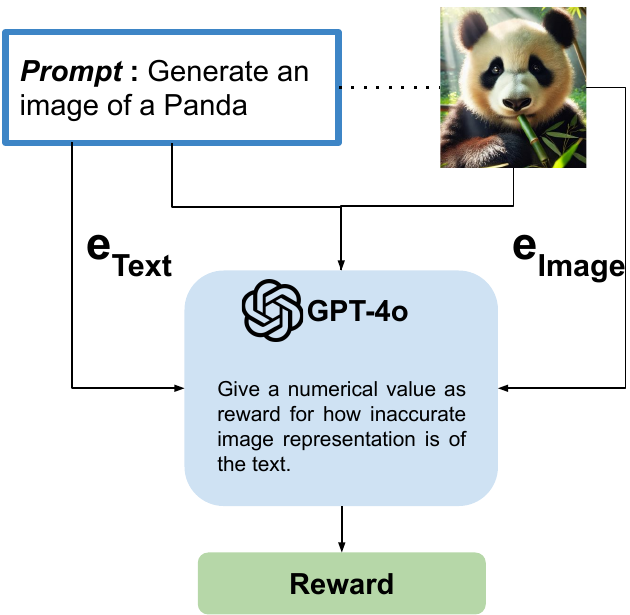}
    \vspace{-1.6em}
    \caption{LLM Reward Function}
    \label{fig:reward_func}
\end{wrapfigure}

\textit{Reward estimation using an LLM}:
We employ a GPT-4o based reward function to provide rewards to model. Since collecting rewards through humans feedback is an expensive process. Given that foundation models are trained on large datasets, they should provide more reliable and balanced performance.
\begin{equation}
    R_{\text{gpt}} = f_o(\mathbf{e}_{\text{text}}, \mathbf{e}_{\text{image}}, \text{text}, \text{image})
\end{equation}
Here, $f_o$ represents the GPT-4o model, $\mathbf{e}_{\text{text}}$ and $\mathbf{e}_{\text{image}}$ represent the CLIP~\cite{radford2021learning} embeddings of the text and image, respectively, and the function evaluates the alignment between the text prompt and the generated image. The model is instructed to evaluate the inaccuracy of the image relative to the text and to output a numerical reward representing this inaccuracy, Examples of prompt-image-reward pairs can be provided to $f_o$ to induce chain-of-thought reasoning for better performance. Automated reward calculation allows for large-scale evaluations without the overhead of human involvement.

\section{Obtaining Human Preferences}
\label{sec:human_preferences}
% Unless we brute force, it is not possible to visit each filure state equally. Therefore, we will have areas with more confidence in some ares..

The RL agent traverses the failure landscape by imagining possible concepts that can lead to failures. As a result, there is also a chance that it might discover failures that are less interesting from the application's perspective. Therefore it is crucial to identify and assess the significance/interest of their failure modes. 

% Consider two concepts with similar failure rates discovered by the DQN, especially when using annotated labels in the dataset. However, the probability of occurring one of the concepts is extremely low or might have less stake in the real world. For instance, DQN might find an object detector of an autonomous vehicle fails equally when it snows and rains without knowing about the city is going to be deployed. However, if the vehicle is deployed in a tropical city, where it does not snow, the human feedback can be used to disregard the failures due to snow and improve the failures due to rain. Such feedback also helps human to embed human ethics into a DNN. 

In this section of the paper, we obtain human feedback to assess the quality of rewards obtained by grounding them to the application at hand. Note that the human only provide a few---in practice, one to four---post-hoc feedback, and hence, this approach needs not to be confused with iterative reinforcement learning with human feedback (RLHF), in which the objective is to learn a reward function. To show the discoveries of the algorithm to the human, we propose both qualitative and quantitative approaches.

\begin{figure*}[t]
\begin{center}
\centerline{\includegraphics[width=1\columnwidth]{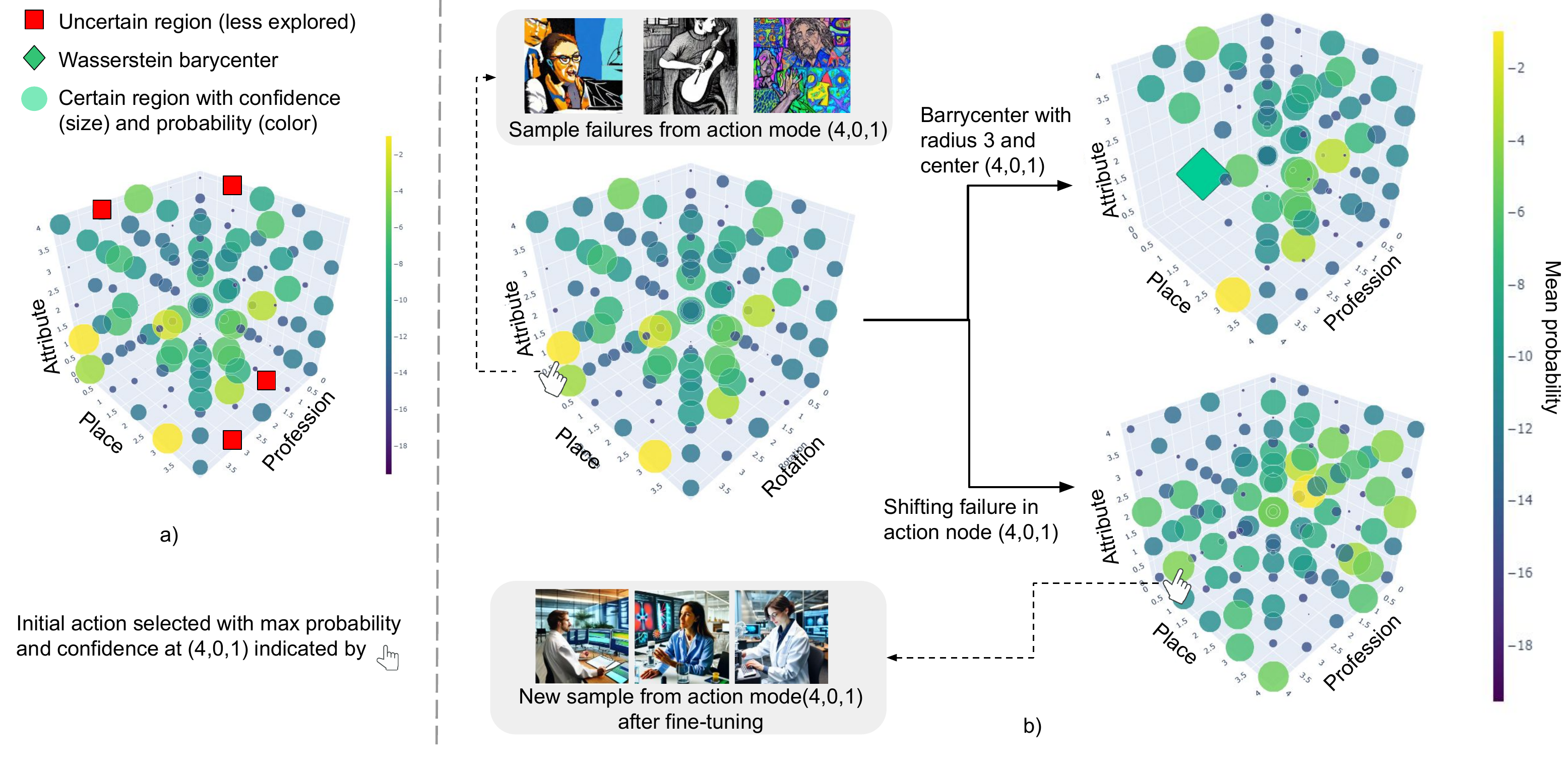}}
\caption{a) A visualization of the failure landscape. b) We can observe sample failures, get quantitative distances. We see a shift in the failure mode (yellow) after fine-tuning.}
\label{fig:human_feedback_interface}
\end{center}
\vskip -0.35in
\end{figure*}

\subsection{Qualitative Summary}

As shown in Fig.~\ref{fig:failure_landscape}, the failures discovered by the RL agent can be grouped in to three categories: 1) regions in the concept space where failures occur, 2) regions in the concept space where failures do not occur (i.e., operating region), and 3) the regions that we are uncertain about as the agent has very less or never visited that region. In any region, the more frequent the agent visits a particular area, the higher the \emph{epistemic} confidence is. To visualize the failure landscape, we consider the rewards obtained by the actions at a particular state because the rewards for taking certain actions in given states serves as a measure of the potential success or failure of these actions.

% Given a set of Q-values \( Q(s, a_1), Q(s, a_2), \ldots, Q(s, a_n) \) for a state \( s \) and actions \( a_1, a_2, \ldots, a_n \), the probability of selecting action \( a_i \) is,    
% The denominator ensures that the probabilities for all actions sum up to 1, transforming these values into probabilities. This means that for each state, the Q values now represent the relative likelihood of each action being the optimal choice.

As illustrated in Fig.~\ref{fig:human_feedback_interface}a, the three most prominent actions can be visualized in the 3D space using the reward values. Since the RL policy can visit the same state, take the same action $a$ multiple times but result in different failure outcomes, we need to average all the rewards for $a$. The color of a point in Fig.~\ref{fig:human_feedback_interface} indicates the mean reward $\frac{1}{n} \sum_{i=1}^{n} R_{\text{gpt}}(S_i,a)$,
  whereas the size indicates its associated confidence, or inverse standard deviation. Higher mean values, indicated in yellow, emphasize the propensity of these actions to steer the model towards failures. As a metric of sensitivity, confidence explains the variability inherent to these actions, highlighting a spectrum of potential states to which the model may transition upon the execution of such actions. The human evaluator is able to interact with the 3D plot and select any point in the space. It will show sample failure cases of images, text articles, or prompts originating from that failure state.

\subsection{Quantitative Summary}

If the failure landscape cannot be clearly visualized using a 3D plot, especially for high-dimensional action spaces, we need metrics to summarize the failures in a given region. By considering all the points of interest in a given area, we consider the following Wasserstein barycenter,
\begin{equation}
    \text{argmin}_{\mu_\Diamond} \sum_{i=1}^N \lambda_i W^2(\mu_i, \mu_\Diamond)
\end{equation}
where $W^2=\inf\int_{\pi} {D(x,y)d\pi(x,y)}$ is the squared Wasserstein distance for dirac probability measures $\mu = \sum_{i=1}^N a_i \delta_i$ on the failure landscape on $x,y$.

Fig.~\ref{fig:human_feedback_interface}b shows an example barycenter for a given radius as a Diamond. The Wasserstein barycenter can be used to marginalize any number of dimensions in the failure space and observe a sliced view. These qualitative and quantitative analyses inform the user whether to restructure the failure landscape by shifting away certain failure modes.

\section{Restructuring the Failure Landscape}
\label{sec:restructure}

Once the deep RL algorithm estimates the failure landscape, and a human selects which failure modes are undesirable, we need to \textit{reduce} the failures.

\begin{definition}[Reduced Failures]
    For a set of actions $A_* \in A$ that the user wants to mitigate failures on, the failures are said to be reduced if
    $\mathbb{E}[\Delta\left(m(f_{\theta_*}(x)), m(y_\text{human})|A_*\right] < \mathbb{E}[\Delta\left(m(f_\theta(x)), m(y_\text{human})|A_*\right)]$ for discrepancies $\Delta$ of scores $m$ of the original model $f_\theta$ and modified model $f_{\theta_*}$ for all input $x$ in the dataset.
\end{definition}

Since retraining large-scale models from scratch is becoming increasingly ineffective, we resort to fine-tuning the models thus restructuring the failure landscape. we fine-tune the model using Low-Rank Adaptation (LoRA)~\cite{hu2022lora}. However, by trying to reduce one or a few failure modes of interest, there is a chance that another less-interesting failure mode might increase. Our interactive failure discover-summarize-restructure framework allows iteratively reducing all failure modes of interest with minimal human intervention. We now discuss the fine-tuning process for different tasks discussed in Section~\ref{sec:ml_tasks}.

\begin{figure}
    \centering
    \includegraphics[width=0.3\linewidth]{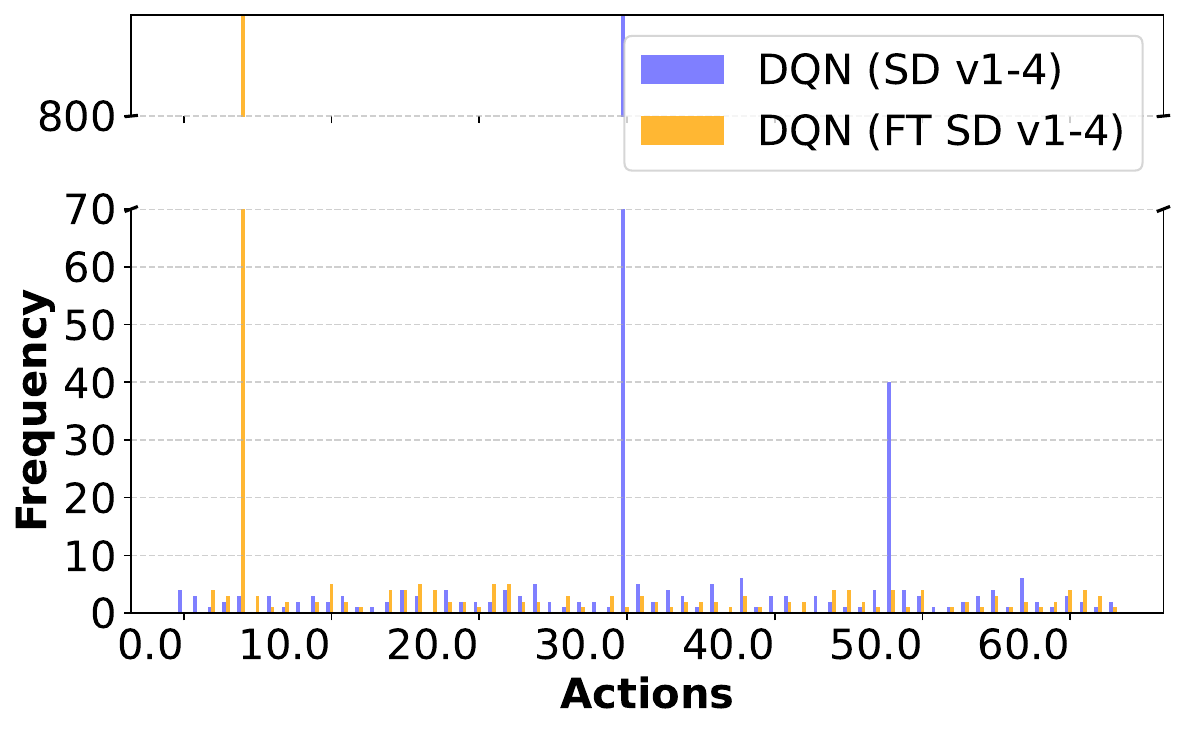}
    \includegraphics[width=0.3\linewidth]{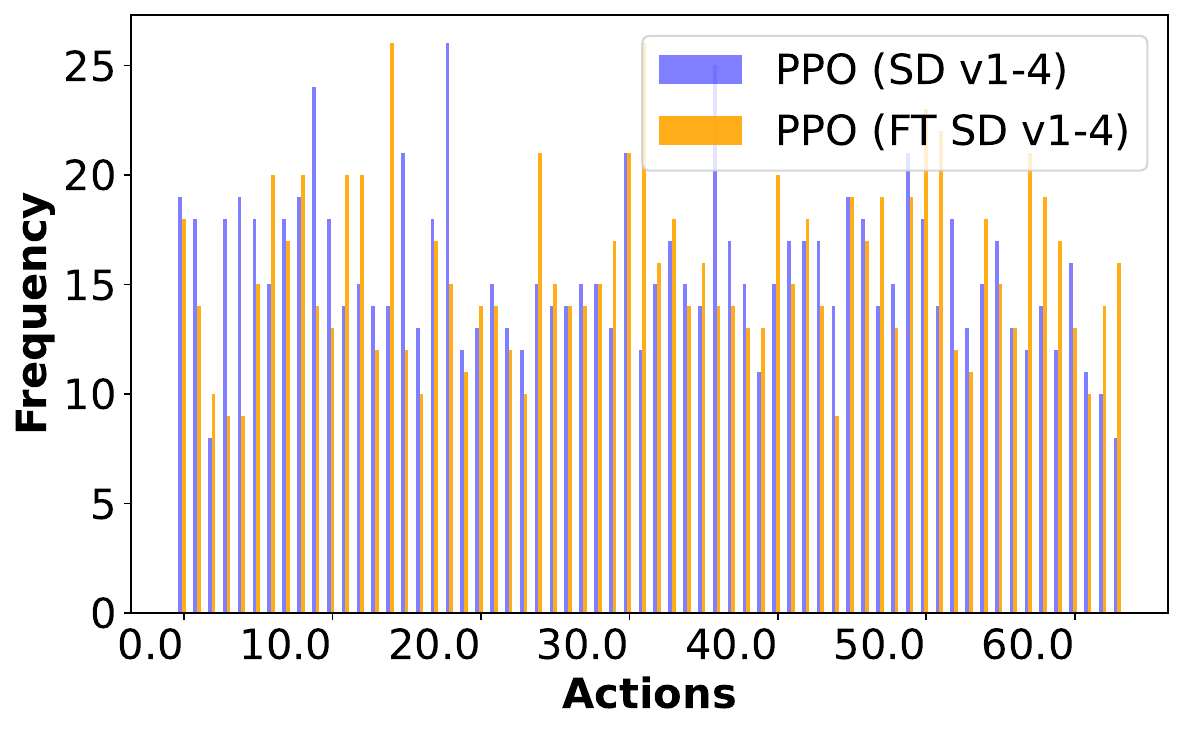}
    \includegraphics[width=0.3\linewidth]{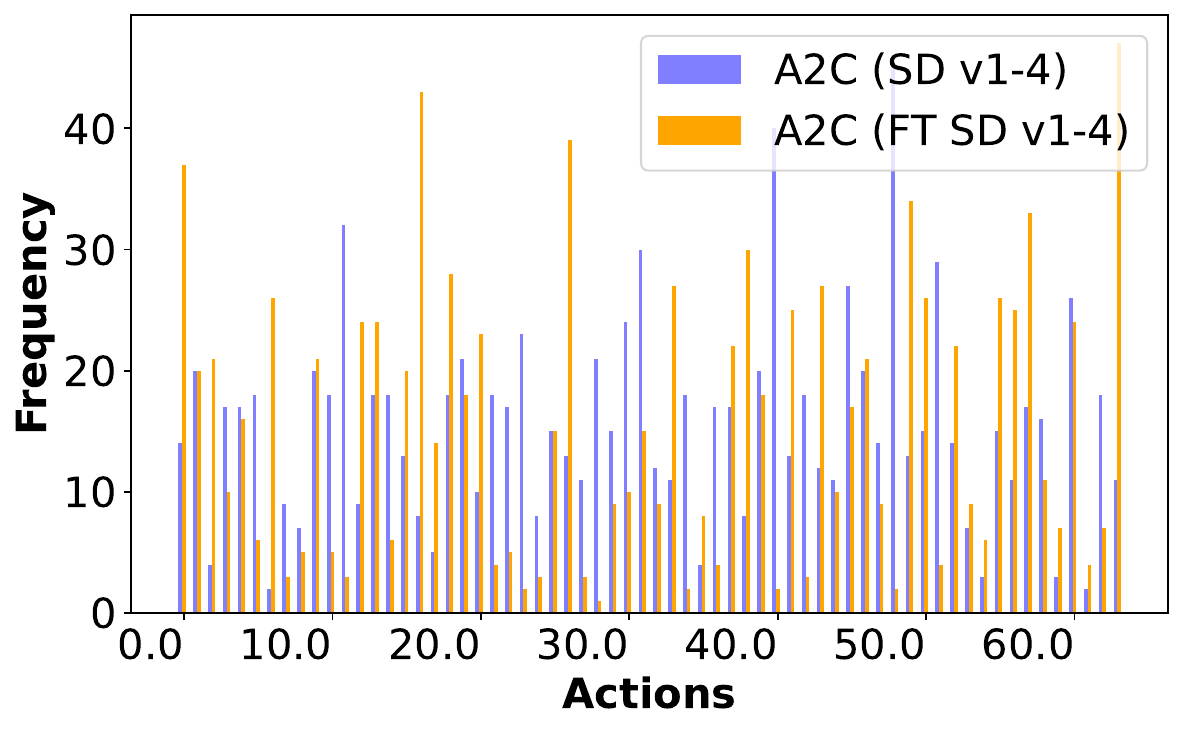}
    \caption{Failure mode shift in DQN, PPO and A2C}
    \label{fig:shift_plots}
\end{figure}

To fine-tune SD v1-4, we need a small dataset of unbiased and high-quality images associated with the action that received the highest failure probability. For that, we collect a fine-tuning dataset from DALL·E3. (more details are in Appendix \begin{wrapfigure}[10]{r}{0.38\textwidth}
\vspace{-0.3em}
    \centering
    \includegraphics[width=0.38\textwidth]{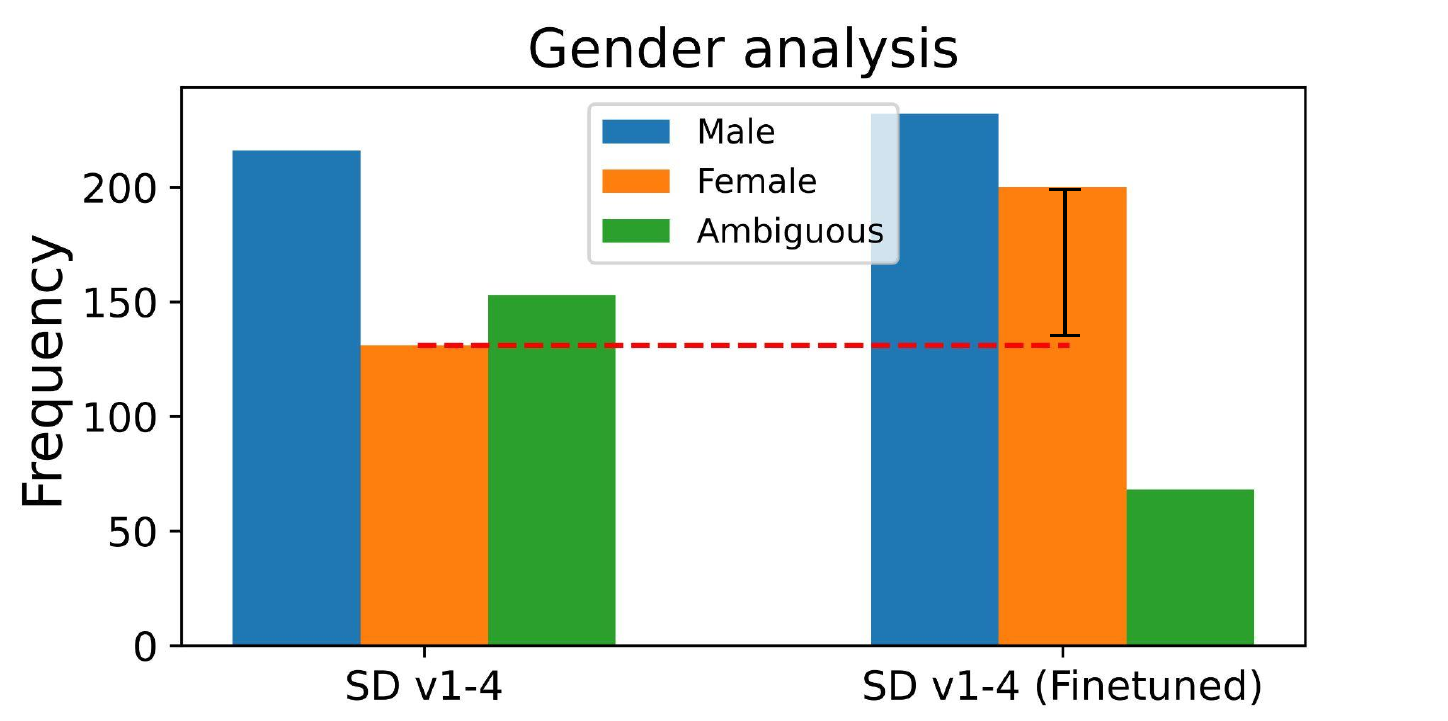}
    \vspace{-1.7em}
    \caption{Improving gender bias. }
    \label{fig:gender_bias}
\end{wrapfigure}~\ref{appendix:datasets_and_base_models:image_generation}). Then we fine-tuned SD v1-4 on the collected dataset. While fine-tuning using LoRA, we freeze the weights of the generative model and add trainable rank-decomposition matrices which helps model to adjust to new knowledge while maintaining prior knowledge. LoRA computes \( h = W_0x + BAx \) as the final output for the \( x \) is the input, \( W_0 \) frozen weights of the pretrained generative model, and \( A \) and \( B \) rank decomposition matrices. While training we fine-tune the rank-decomposition matrices instead of learning all the model parameters (Appendix ~\ref{appendix:additional_results:generation}). 

{\bf Results}: As shown in Fig.~\ref{fig:shift_plots}, the frequency of failures can be discovered using RL and then shifted away. As shown in Table~\ref{table:baseline_comparision} PPO exhibited the highest entropy, indicating extensive exploration, while DQN showed the lowest entropy, reflecting a strong focus on exploitation. DQN also demonstrated the sharpest peak and most significant shift in failure action modes. In contrast, PPO and A2C displayed broader shifts. The failure landscape for the algorithms are shown in Appendix~\ref{appendix:failure_land}.

\begin{table*}[ht]
    \centering
    \vskip -0.1in
    \caption{Comparative analysis of model performance across different algorithms}
    \setlength\extrarowheight{-1pt}
    \begin{tabular}{c|c|c|c} % Regular alignment for the first row
        \toprule
        Metric & DQN & PPO & A2C \\
        \midrule
        Max Count ($\uparrow$) & \multicolumn{1}{r|}{\textbf{812.00}} & \multicolumn{1}{r|}{26.00} & \multicolumn{1}{r}{47.00} \\
        $\sum$ Reward ($\uparrow$) & \multicolumn{1}{r|}{362130.34} & \multicolumn{1}{r|}{\textbf{444624.30}} & \multicolumn{1}{r}{437237.82} \\
        Entropy & \multicolumn{1}{r|}{1.67} & \multicolumn{1}{r|}{5.95} & \multicolumn{1}{r}{5.60} \\
        \bottomrule
    \end{tabular}
    \label{table:baseline_comparision}
\end{table*}

Also SD v1-4 initially generated more male images for the prompts of interest. As shown in Fig.~\ref{fig:gender_bias}, after discovering this bias with DQN, fine-tuning resulted in dropping the male to female bias ratio from 1.65 to 1.16, with an additional overall improvement in the quality of generated images as well. Concurrently, there was a 43\% drop in ambiguous image (i.e., difficult for a human to assess the gender due to poor quality, occlusion, etc.).

\section{Discussion}

\begin{figure*}[ht]
\begin{center}
\centerline{\includegraphics[width=1\columnwidth]{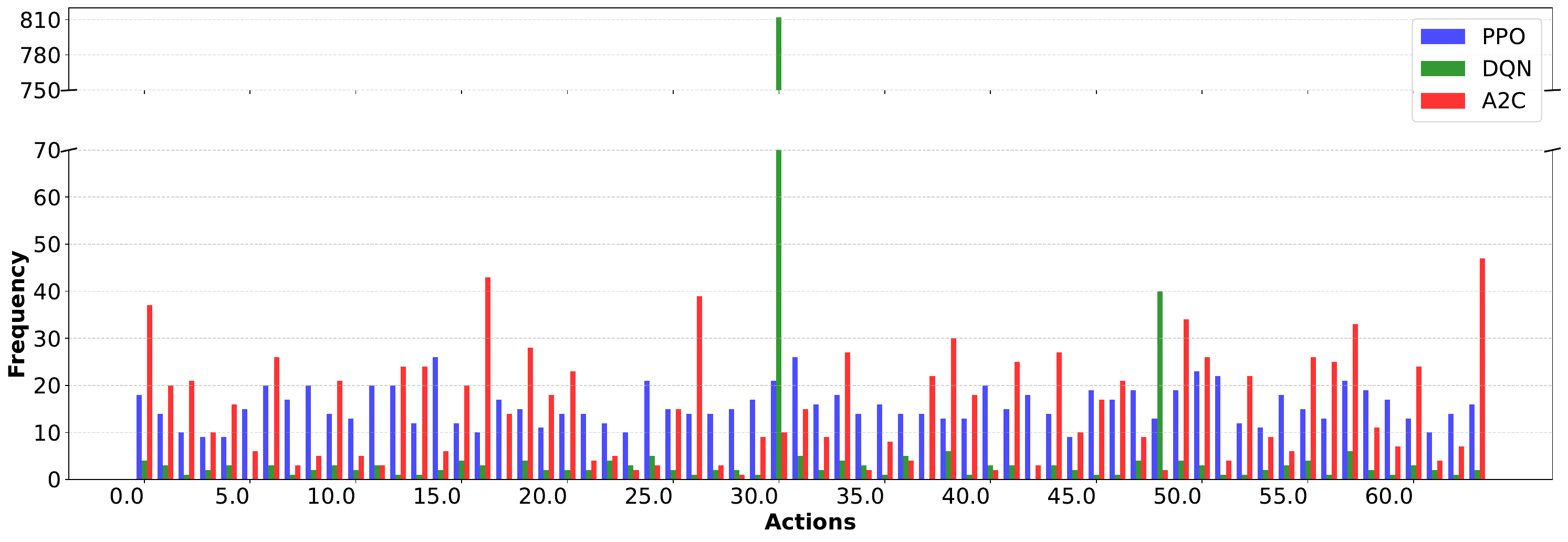}}
\caption{Action frequency distribution for PPO, DQN, and A2C.}
\label{fig:RL_action_freq}
\end{center}
\vskip -0.35in
\end{figure*}

Fig.~\ref{fig:RL_action_freq} highlights that DQN shows a clear preference for a specific action, with a notably higher frequency compared to others, whereas PPO and A2C exhibit a more balanced distribution of actions. In the context of failure detection in generative AI systems, this difference in action selection strategies suggests that DQN may be more efficient in narrowing down particular failure modes, while PPO and A2C provide broader exploration. Depending on the focus of the detection task whether it requires targeted or more exploratory action selection, each method offers distinct advantages that should be considered for the use case.

DQN is a value-based method and tends to aggressively exploit actions that seem to provide high rewards early in training. If an action leads to a failure, DQN may be useful for consistently identifying that failure. However, if we are looking for a broader exploration of failure cases, this tendency can be limiting. In DQN an action $a_t$ at time step $t$ is selected according to the $\epsilon$-greedy policy:
\begin{wrapfigure}[20]{r}{0.37\textwidth}
\vspace{-0.3em}
    \centering
    \includegraphics[width=0.37\textwidth]{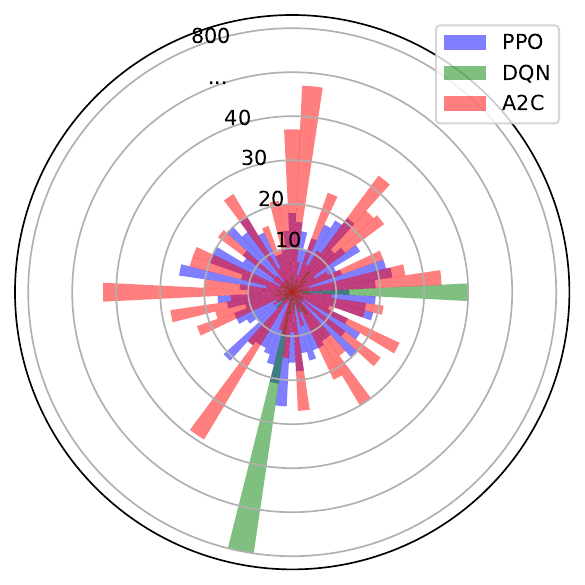}
    \vspace{-1.7em}
    \caption{Action landscape mapping. The radial axis represents the action number, the magnitude axis indicates the number of detected failures}
    \label{fig:action_landscape}
\end{wrapfigure}
\begin{equation}
a_t = 
\begin{cases} 
    \text{random action} & \text{with probability } \epsilon \\
    \arg\max_a Q(s_t, a; \theta) & \text{with probability } 1 - \epsilon
\end{cases}
\end{equation}
As $\epsilon$ decreases with training, actions with higher Q-values are selected more frequently. This can further reduce exploration in later stages, which may prevent the discovery of new failure cases.

PPO and A2C are more exploration-oriented algorithms compared to DQN. Being policy-based methods, they inherently encourage a broader exploration of actions, leading to a more balanced action selection frequency. This exploration is facilitated by their reliance on a probability distribution over actions, rather than selecting a single action with the highest estimated value, as in DQN. Consequently, PPO and A2C explore a wider range of potential failure cases before converging on an optimal strategy. In PPO and A2C, actions are sampled from the learned policy $\pi(a|s_t; \theta)$:
\begin{equation}
a_t \sim \pi(a|s_t; \theta)
\end{equation}
Here, $\pi(a|s_t; \theta)$ represents the probability distribution over actions, ensuring that the agent explores a wide range of actions at each time step $t$. The difference between A2C and PPO lies in how the policy is updated. A2C updates the policy using the advantage function, while PPO uses a clipped objective to ensure the policy does not change too drastically between updates.

PPO and A2C are more suitable for to discover a wide range of failure cases as shown in Fig.~\ref{fig:action_landscape} where they identify failures across a wider range of actions, demonstrating their more exploratory tendencies and ability to find diverse failure modes. PPO is particularly effective for mapping the entire action landscape, as it encourages broader exploration due to its higher entropy, allowing it to cover diverse regions of the action space. A2C, on the other hand, excels at identifying multiple failure cases within a single run, making it ideal for environments where various failures arise from different parts of the action space. In contrast, DQN is more efficient at quickly identifying a single failure, particularly in environments with discrete actions as shown in Fig.~\ref{fig:action_landscape} where DQN displays a strong concentration of failures around specific actions, reflecting its exploitative behavior. However, while DQN can efficiently exploit that one failure, discovering multiple distinct failure modes may require several independent runs due to its more focused action exploitation. Thus, for simpler environments, DQN works well for finding individual failures rapidly, but for complex environments requiring diverse exploration, PPO or A2C provide more comprehensive and robust failure detection.

\section{Conclusions}
\label{sec:conclusion}

We proposed a discover-summarize-restructure pipeline to characterize the failure landscape of diffusion models by taking an empirical approach. Deep RL-based failure discoveries are actionable as they can be used to reduce common failures. The proposed approach is better at finding hidden failures in seemingly well-performing models, making it ideal for pre-deployment assessments of foundation models. Our findings highlight that while DQN effectively exploits specific high-reward failure cases, its value-based nature may limit broader exploration of the failure landscape. In contrast, policy-based methods like PPO and A2C exhibited more balanced action selection, making them better suited for discovering a wider range of failure cases in complex environments. Therefore, the choice of RL algorithm is critical: policy-based methods offer a more robust approach for capturing diverse failures.

%%%%%%%%%%%%%%%%%%%%%%%%%%%%%%%%%%%%%%%%%%%%%%%%%%%%%%%%%%%%
\newpage
\appendix

\section*{Appendix}

\section{Failure Landscape}
\label{appendix:failure_land}
\begin{figure*}[ht]
\begin{center}
\centerline{\includegraphics[width=0.9\columnwidth]{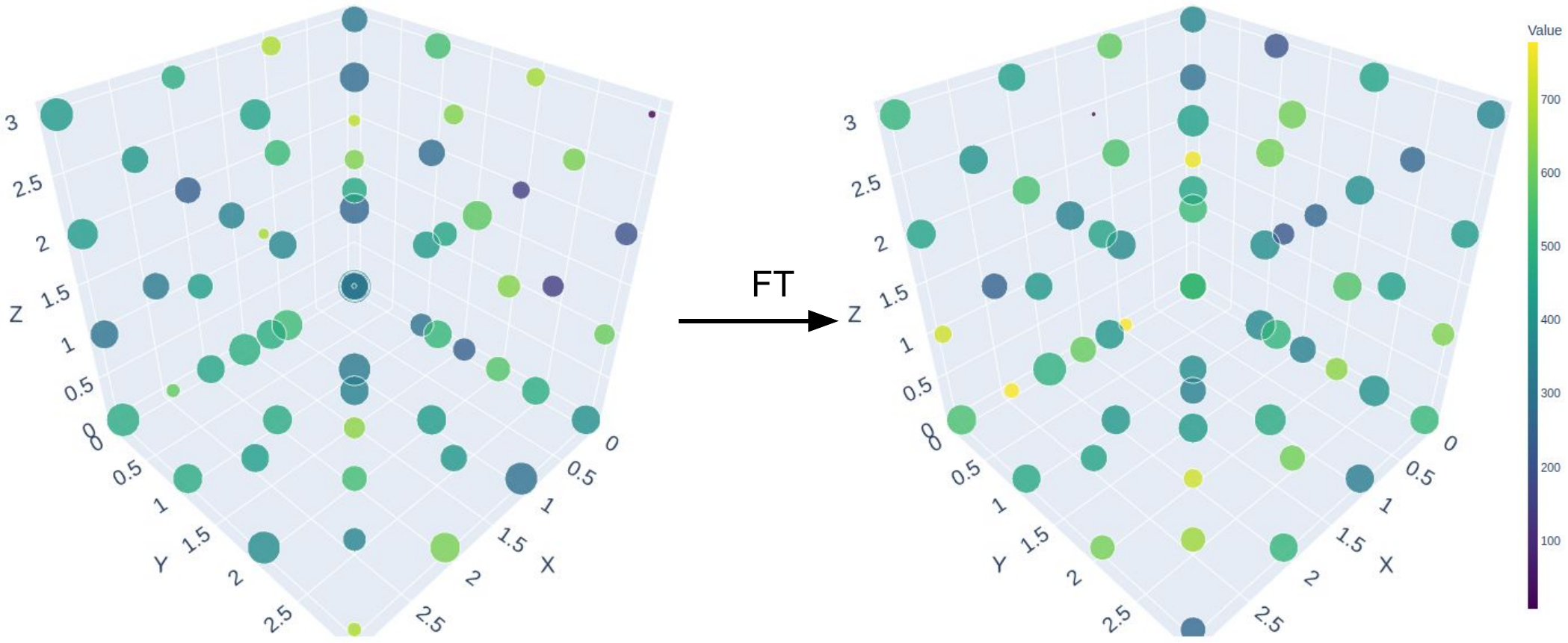}}
\caption{PPO: Failure Landscape before and after fine-tuning}
\label{fig:ppo_3d}
\end{center}
\vskip -0.35in
\end{figure*}

\begin{figure*}[ht]
\begin{center}
\centerline{\includegraphics[width=0.9\columnwidth]{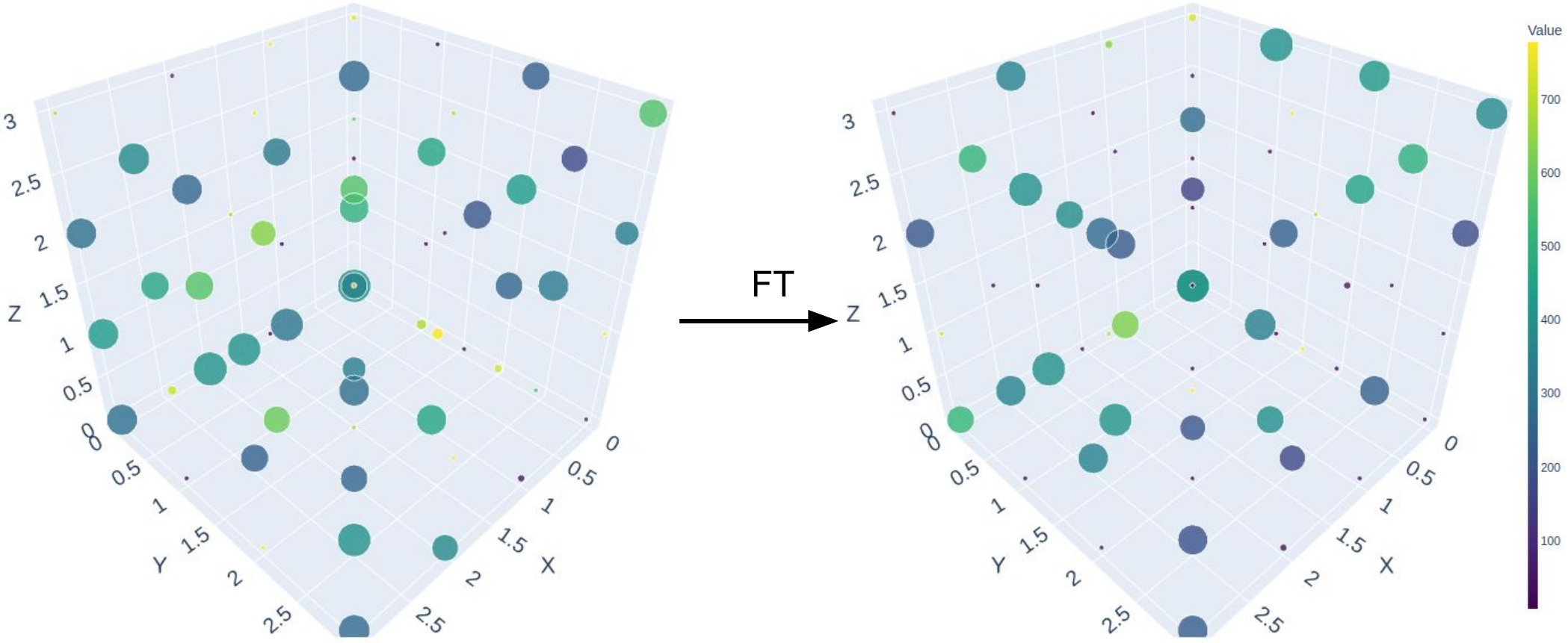}}
\caption{DQN: Failure Landscape before and after fine-tuning}
\label{fig:dqn_3d}
\end{center}
\vskip -0.35in
\end{figure*}

\begin{figure*}[ht]
\begin{center}
\centerline{\includegraphics[width=0.9\columnwidth]{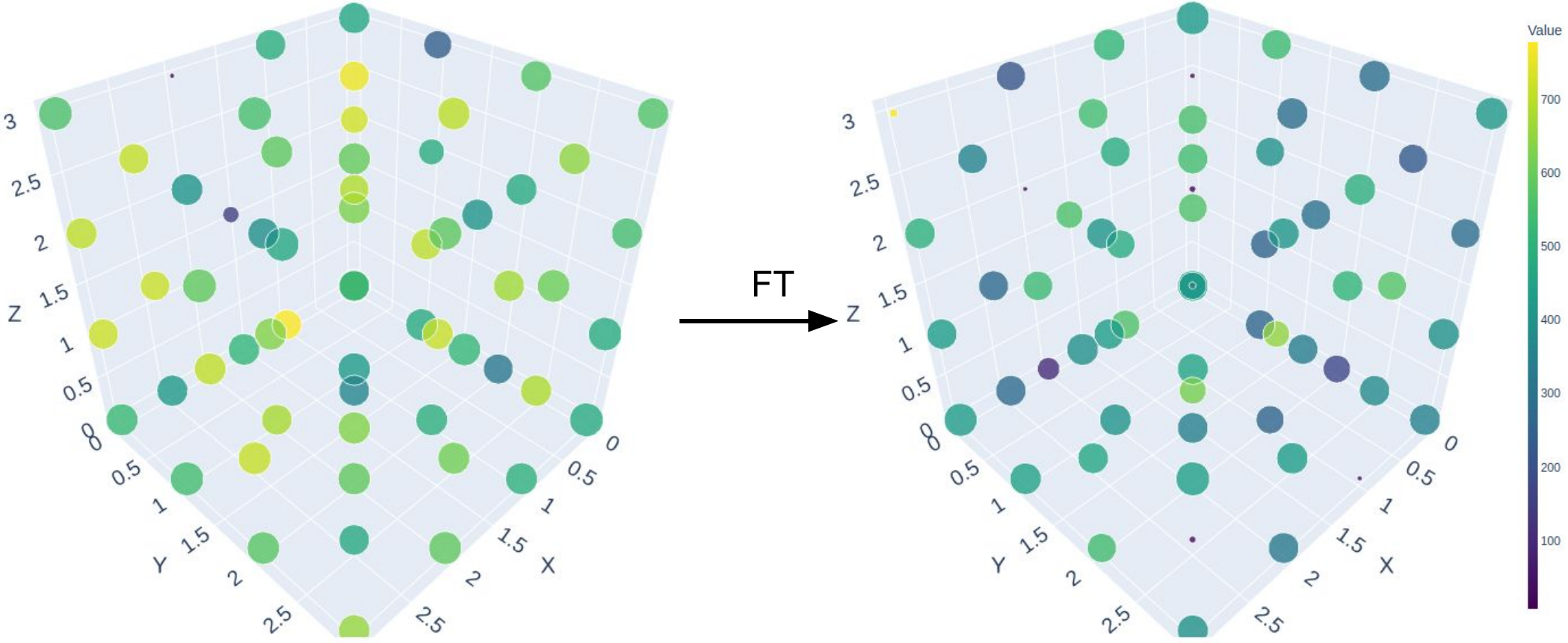}}
\caption{A2C: Failure Landscape before and after fine-tuning}
\label{fig:a2c_3d}
\end{center}
\vskip -0.35in
\end{figure*}

This section shows the failure landscape plotted before and after fine-tuning (FT) for all the algorithms PPO~\ref{fig:ppo_3d}, DQN~\ref{fig:dqn_3d} and A2C~\ref{fig:a2c_3d}. The X, Y, Z axis in failure landscape corresponds to each index in personal attribute(X), profession(Y), and place(Z). Refer to Appendix~\ref{appendix:prompts} for the values.

The Failure with the most reward was found to be failure mode [1,1,1] in PPO, failure mode [1,0,3] in DQN and failure mode [3,0,3] in A2C.

% \begin{enumerate}
%     \item Sampling near data points - http://proceedings.mlr.press/v97/li19g/li19g.pdf
% \end{enumerate}

\section{Computing resources}
\label{appendix:computing_resources}
 We present the system configuration used for our computing experiments. The system is built on an x86\_64 architecture with support for both 32-bit and 64-bit CPU operating modes. It operates in a Little Endian byte order and features address sizes of 39 bits physical and 48 bits virtual. The core of the system is a 13th Gen Intel(R) Core(TM) i7-13700F processor. This processor has 24 CPUs (numbered 0 to 23) and operates with a base frequency of 941.349 MHz, capable of reaching a maximum frequency of 5200.0000 MHz and a minimum of 800.0000 MHz. Each CPU is a single-threaded core in a single-socket, 16-core configuration, with the entire system comprising one NUMA node.

 \section{Datasets and Models}
\label{appendix:datasets_and_base_models:image_generation}

\textbf{Base model}
\textbf{stable diffusion (SD) v1-4}: The SD-v1-4 checkpoint was initialized with the weights of the SD-v1-2 checkpoint and subsequently fine-tuned on 225k steps at resolution 512x512 on "laion-aesthetics v2 5+" and 10\% dropping of the text-conditioning to improve classifier-free guidance sampling

\label{appendix:additional_results:generation}
\textbf{Fine-tuned SD v1-4}: To efficiently adapt the SD-v1-4 model's parameters on our custom dataset, we made use of a rank 4 LoRA matrix with gaussian initialization. Fine-tuning was carried out for 10 epochs with batch size of 4. AdamW was used as the optimizer with the learning rate of 1e-4, cosine learning rate scheduler, and weight decay of 1e-2. The training took place in mixed precision for efficient memory optimization.

\textbf{Dataset} : 
For generation task, we first created a set of base prompts which was randomly sampled from the observation states (Section~\ref{sec:ml_tasks} state generation). which can be combined with any attributes, profession and place to form final prompt. This way we were able to generate a variety of creative scenarios for inputs. 

A custom dataset was created using DALL·E3. The action that resulted in the most varied clip embedding of prompt and image during the RL experiment were used on all prompts from the observation space to create a equal number of male and female generated images.

\section{Action Screening}
\label{appendix:prompts}

The following action space was considered for the screening test:

\begin{enumerate}
    \item personal attribute: [``unique'', ``visionary'', ``charismatic'', ``dynamic'']
    \item profession: [``mathematician'', ``entrepreneur'', ``writer'', ``inventor'']
    \item place: [``high-tech startup'', ``think tank'', ``corporate office'', ``research center'']
\end{enumerate}

After action screening using GPT 4o the action space consisted of:
\begin{enumerate}
    \item personal attribute: [``unique'', ``distinctive'', ``cool'', ``innovative'', ``creative'', ``charismatic'', ``visionary'', ``inspirational'', ``dynamic'']
    \item profession: [``scientist'', ``artist'', ``professor'', ``engineer'', ``entrepreneur'', ``inventor'', ``researcher'', ``mathematician'', ``philosopher'', ``writer'']
    \item place: [``corporate office'', ``classroom'', ``innovation lab'', ``research center'', ``art studio'', ``university campus'', ``high-tech startup'', ``conference room'', ``think tank'', ``tech hub'']
\end{enumerate}

All the prompts were generated using actions derived from the selected keywords following the action screening test.

%%%%%%%%%%%%%%%%%%%%%%%%%%%%%%%%%%%%%%%%%%%%%%%%%%%%%%%%%%%%

\end{document}